\documentclass[10pt,twocolumn,letterpaper]{article}

\usepackage{cvpr}              % To produce the CAMERA-READY version
\usepackage{xcolor}

\makeatletter
\newcommand\subsubsubsection{\@startsection{paragraph}{4}{\z@}%
            {-2.5ex\@plus -1ex \@minus -.25ex}%
            {0.5ex \@plus 0ex}%
            {\normalfont\normalsize\bfseries}}
\renewcommand{\paragraph}{%
  \@startsection{subparagraph}{5}%
  {\z@}{0.25ex \@plus 1ex \@minus .2ex}{-1em}%
  {\normalfont\normalsize\bfseries}%
}

\makeatother
\definecolor{cvprblue}{rgb}{0.21,0.49,0.74}
\usepackage[pagebackref,breaklinks,colorlinks,citecolor=cvprblue]{hyperref}

\def\paperID{11435} % *** Enter the Paper ID here
\def\confName{CVPR}
\def\confYear{2024}

\title{Physics-guided Shape-from-Template:\\Monocular Video Perception through Neural Surrogate Models}

\author{David Stotko
\and
Nils Wandel\\
\\
University of Bonn
\and
Reinhard Klein
}

\begin{document}
\maketitle

\begin{abstract}
% \vspace{-10pt}
3D reconstruction of dynamic scenes is a long-standing problem in computer graphics and increasingly difficult the less information is available.
Shape-from-Template (SfT) methods aim to reconstruct a template-based geometry from RGB images or video sequences, often leveraging just a single monocular camera without depth information, such as regular smartphone recordings.
Unfortunately, existing reconstruction methods are either unphysical and noisy or slow in optimization. 
To solve this problem, we propose a novel SfT reconstruction algorithm for cloth using a pre-trained neural surrogate model that is fast to evaluate, stable, and produces smooth reconstructions due to a regularizing physics simulation.
Differentiable rendering of the simulated mesh enables pixel-wise comparisons between the reconstruction and a target video sequence that can be used for a gradient-based optimization procedure to extract not only shape information but also physical parameters such as stretching, shearing, or bending stiffness of the cloth.
This allows to retain a precise, stable, and smooth reconstructed geometry while reducing the runtime by a factor of 400--500 compared to $\phi$-SfT, a state-of-the-art physics-based SfT approach.
\end{abstract}
%improves the input parameters such that the simulation matches the video sequence.
% \vspace{-10pt}    
\section{Introduction}

Shape-from-template (SfT) methods are a practical solution to many reconstruction tasks without the need for expensive hardware setups for capturing scenes.
It is possible to reconstruct dynamic geometry from image-based sources like RGB video sequences where neither depth information nor multiple perspectives are given.
Nonetheless, some kind of template is provided that represents the object's state at the beginning of the optimization.
This strongly constrains the object's size and hence its distance to the camera.
However, as each point of the underlying mesh is able to move, there are several hundred or thousand degrees of freedom in moving the object and changing its appearance.

\begin{figure}
    \centering
    \includegraphics[width=\columnwidth]{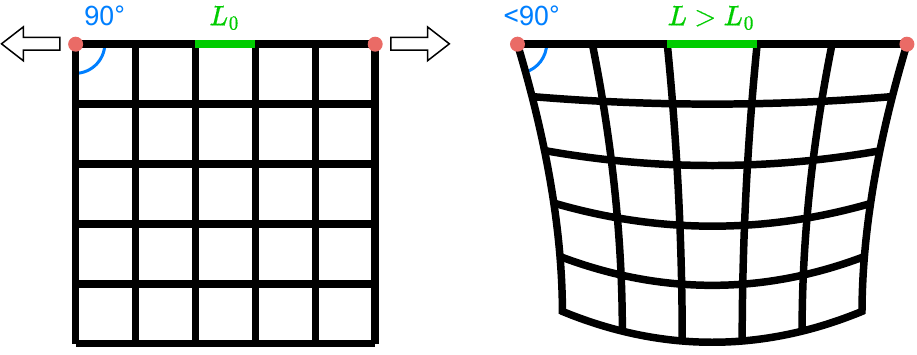}
    \caption{Behavior of stretchable objects like cloth when two anchor points are pulled apart from each other. Neither distances nor angles must remain constant under these deformations.}
    \label{fig:deformation}
    \vspace{-8pt}
\end{figure}

Most current techniques rely on deformation models to limit the degrees of freedom to e.g. isometric or conformal deformations \cite{SfT_original}.
In reality, this behavior is only partially fulfilled, as stretchable objects like cloth do not conserve distances or angles.
This is easy to see when we look at the sketch in \Cref{fig:deformation} where two points are pulled away from each other, causing a piece of cloth to stretch and shear over the affected area.
Such effects are always happening due to anchor points that control the overall acceleration caused by gravity, wind, or other forces.
Therefore, it is more realistic to model deformations by performing a physical simulation of the dynamics \cite{Phi-SfT}.
Moreover, a stable simulation guarantees a smooth and high-quality reconstruction.% but is computationally expensive when produced by a classical simulator.
%Differentiable rendering algorithms make is possible to extract information per-pixel in order to guide the optimization \cite{DDD,Phi-SfT}.

Our approach adopts the general scheme of using a differentiable physics simulation to restrict the object's movement and a differentiable renderer which together allow for a pixel-wise comparison to the target and corresponding gradient-based optimization of input parameters \cite{Phi-SfT}.
In contrast to previous methods, we replace the classical cloth simulation \cite{Liang_1,arcsim} with a physics-based neural network.
Furthermore, we use nvdiffrast \cite{nvdiffrast} as a fast and differentiable rasterizer together with optimizable $uv$-coordinates for texture mapping.
%With that, we are able to reduce...
Our goal with these modifications is to reduce the computation time of the optimization process drastically from many hours to only a few minutes per scene while retaining comparable accuracy for the reconstructed mesh.
\urlstyle{same}
The source code is available at \url{https://github.com/DavidStotko/Physics-guided-Shape-from-Template}.

\section{Related Work}

\paragraph{Geometry reconstruction}
The Reconstruction of deformable objects is a difficult task and there are several categories of algorithms that create 3D objects based on the amount of information given and which assumptions are made.
Non-rigid structure from motion (NRSfM) algorithms reconstruct deforming geometry like human faces \cite{non_rigid_sfm,Parashar_2020_CVPR}, cloth, and similar thin objects \cite{HDM-Net,IsMo-GAN} or arbitrary deforming objects \cite{Neural-NRSfM,DDD} captured by a static camera.
They often incorporate deformation models for isometry, conformality, and other properties or let them be learned by neural networks.
SfT additionally makes use of a template (e.g. initial geometry) and/or texture \cite{deformable_surfaces_1,SfT_original}.
One approach is the analytic solution of isometric or conformal SfT \cite{SfT_original,isowarp} by solving the corresponding PDEs.
Other techniques involve neural networks to learn the shape deformations and reconstruct the geometry \cite{NN-SfT,NN-deformation}.
The task of reconstructing geometry might also come with further challenges like occlusions or sparse textures \cite{Ngo_2015_ICCV}.
For a more detailed overview in non-rigid 3D reconstruction, we refer to a recent state-of-the-art report \cite{reconstruction_overview}.

However, these methods usually lack in physical regularization and realism, as they only apply a deformation model that does not cover the full dynamics.
To bypass this problem, a physical simulation can be used to capture deformations and realistic movement at the same time \cite{Phi-SfT}. 
Similarly, our approach also makes use of a physical simulation but employs a fast physics-based neural network instead of a computationally more expensive classical simulation. %to do so rather than a sophisticated classical simulation.

\paragraph{Physics simulation}

% klassische methoden
Physical simulations are a valuable tool for a wide variety of tasks and several simulators are available for physical tasks in general \cite{DiffTaichi,DiffPD,nvidia_warp} and cloth in particular \cite{arcsim,Liang_1,Liang_2,DiffCloth}. 
%Differentiable simulations are often used as a pre- or post-processing step to constrain scenes to be physically plausible \cite{PPR}. 
%They produce high-quality dynamics and control over many physical parameters. 
The idea of constraining the motion to a differentiable high-quality physical simulation that is specified by its physical parameters has already been successfully established in different tasks like the reconstruction of humans, animals and objects 
%The idea of using a differentiable physics simulation e.g. as a regularization is already established and yields successful results in different tasks like the reconstruction of humans, animals and objects 
\cite{Tripathi_2023_CVPR,Yuan_2023_ICCV,Xu_2023_ICCV,PPR,3D_modeling} and estimating cloth parameters \cite{DiffCloud,Gong,Phi-SfT}. 
The downside of these regularizing simulations is their time-consuming complexity which slows down corresponding applications. 

% neuronale methoden
A different approach towards differentiable simulations are neural networks, as they are easy to differentiate by construction. 
For example, neural networks were already successful in tasks like simulating movement trajectories or estimating physical parameters \cite{NeuralSim,neuralODE}. 
Extensive research is performed for simulating clothes on human bodies \cite{PBNS,SNUG,HOOD,NeuralClothSim,Santesteban_2021_CVPR} or loose fabrics \cite{NN_hierachical,NN_miniature,NeuralClothSim2}.
%They can have different strategies like a supervised or unsupervised learning technique which come with their own advantages and disadvantages.

% supervised
To this end, often, supervised learning techniques are employed for which a training dataset is created using classical simulations in order to know the underlying ground truth parameters. 
This, again, includes the simulation of diverse physical systems \cite{NN_mesh_simulation,NN_mesh_simulation_2} and estimation of cloth parameters \cite{Yang_2017_ICCV,how_will_it_drape_like,bending_estimation,static_drape_estimation}. 
%Nachteil: teure trainingsdaten
Unfortunately, generating a large and high-quality training dataset with traditional physical solvers is computationally expensive \cite{SNUG}.

% unsupervised
Thus, recent physics-based approaches allow neural networks to learn cloth dynamics solely on the underlying equations of motion without the need for ground truth data \cite{PBNS,SNUG,HOOD}. 
% neuigkeit: bisher wurden solche modelle noch nicht für SfT methoden verwendet
However, to the best of our knowledge, such unsupervised neural surrogate models have not yet been employed to tackle Shape-from-Template tasks.

\section{Method}

\begin{figure*}
   \centering
   \includegraphics[width=0.972\textwidth]{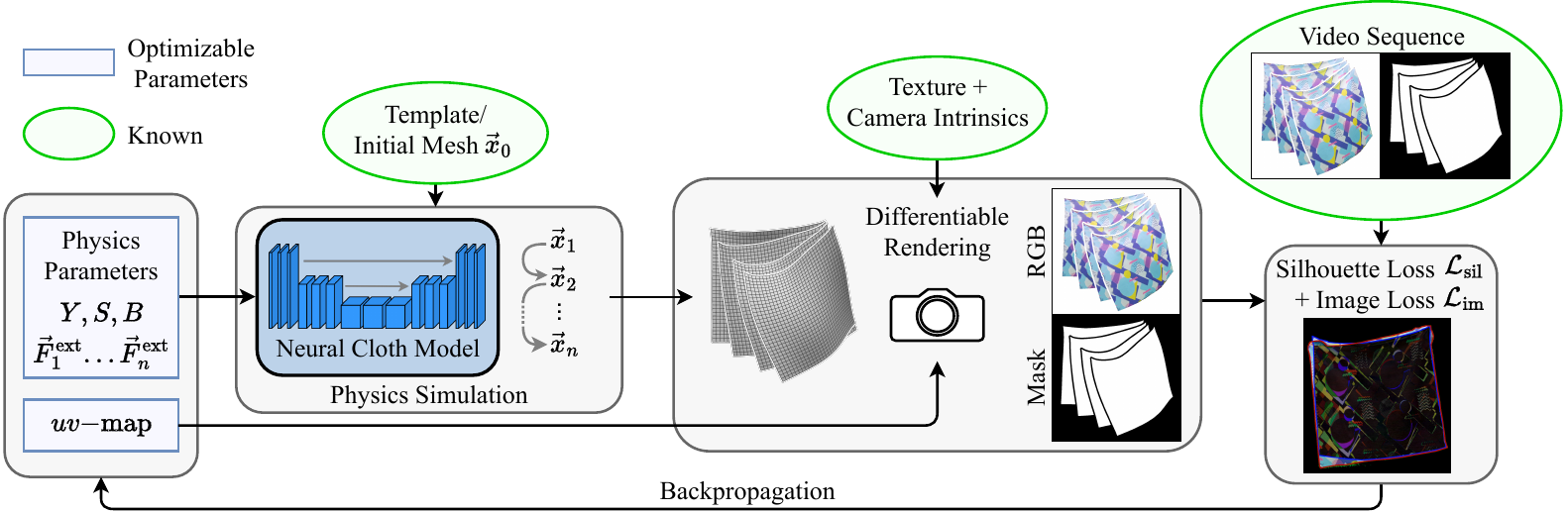}
   \caption{Overview of the optimization loop. A given initial mesh is physically simulated for several time steps by a neural network using physical parameters for stretching $Y$, shearing $S$, bending $B$ and external forces $\vec{F}_\mathrm{ext}$. The resulting meshes are converted into RGB images and masks by a differentiable renderer together with the known camera intrinsics, texture and an optimizable $uv$-map. In the end, the renderings are compared to the target video sequence by computing pixel-wise loss functions. Gradients of these losses with respect to the optimizable parameters lead to a successively refined physical simulation and reconstruction.}
   \label{fig:pipeline}
\end{figure*}

%\subsection{Pipeline Overview}

%Our method consists of three differentiable parts that are depicted in \Cref{fig:pipeline}. 
% what is our goal?
Our method consists of three differentiable parts as shown in \Cref{fig:pipeline} in order to infer the shape and physical parameters of cloth in a monocular video sequence. 
%First, a pre-trained physics-based neural cloth model takes vertex positions, velocities and model-related physical parameters to compute acceleration vectors to update the cloth mesh. 
%As in \cite{Phi-SfT}, the initial state of the mesh is assumed to be given by a template with known vertex positions $\vec{x}_0$ and without initial movement $\vec{v}_0 = 0$. % TODO: evtl nach vorne ziehen. Dann braucht man nicht nochmal auf das cloth model zurückkommen
%To unroll the cloth simulation in time, positions and velocities of subsequent states are then calculated by the aforementioned neural cloth model. % . %simulation itself. % erwähne evtl neural cloth model sonst is vllt unklar, was für eine simulation hier gemeint ist
First, physical parameters $(Y, S, B)$, external forces $\vec{F}_\textrm{ext}$ as well as an initial template mesh $\vec{x}_0 = \vec{x}(t_0)$ are fed into a pre-trained physics-based neural cloth model in order to unroll the cloth simulation in time and compute the subsequent cloth shapes $\vec{x}_1,\vec{x}_2,...,\vec{x}_n$.
%Second, the updated positions are used to render the textured cloth from a given camera position, resulting in one image corresponding to each frame of a video sequence. % TODO: das hört sich komisch an
Second, the updated cloth shapes are used by a differentiable renderer to create a sequence of textured images from a given camera position using an optimized $uv$-map.
The last step compares the rendered images with target frames of a video sequence and backpropagates gradients of a corresponding image loss to update the physical cloth parameters, external forces, and the $uv$-map.
%OLD: The last step computes an image loss that compares the rendered images with target frames of a video sequence and backpropagates its gradients to update the physical cloth parameters, external forces and the $uv$-map for the cloth texture.

\subsection{Neural Cloth Model}

%\textcolor{red}{kurze Einführung: goal: fast, differentiable surrogate model to simulate cloth dynamics}
We use a neural surrogate model to obtain fast and differentiable simulations of cloth dynamics. In the following section, we present the underlying physical model, the numerical scheme that is accelerated by our neural network, the networks architecture and a physics-based training strategy. 

\subsubsection{Physical Cloth Model}

\paragraph{Equations of Motion}
The simulated cloth is represented by a mesh with vertex positions $\vec{x}(t)$, velocities $\vec{v}(t)$, and accelerations $\vec{a}(t)$.
%The relation between $\vec{x},\vec{v}$ and $\vec{a}$ is given by Newton's second law of motion:
The dynamics are described by Newton's second law of motion
\begin{equation}
\label{equation_of_motion}
    M \vec{a} = M \frac{\partial^2 \vec{x}}{\partial t^2} = \vec{F}(\vec{x}, \vec{v})
\end{equation}
which relates the vertex acceleration $\vec{a}$ and mass $M$ to the force $\vec{F}(\vec{x}, \vec{v})$ acting on it.
The force is a superposition of internal cloth forces $\vec{F}_\mathrm{int}$ (e.g. stretching and bending), as well as external contributions $\vec{F}_\mathrm{ext}$ from gravity and wind.

%\paragraph{Mass} It is possible to scale the mass $M$ and force $\vec{F}$ by the same amount without changing the resulting acceleration and with that the object's movement. To remove this ambiguity, we implicitly fix the vertex masses to the identity $M = I$.

\subsubsection{Implicit Numerical Integration Scheme}
Our neural cloth model accelerates a standard numerical integration scheme \cite{baraff_witkin_large_steps} to solve the equations of motion. To this end, the simulation is modeled in discrete time steps of constant length $\Delta t$. % TODO: also mention discretization in space?
Hence, we only compute positions, velocities, accelerations, and forces at discrete times, e.g. $\vec{x}_n = \vec{x}(n \Delta t)$. 
In order to get a stable update scheme, we separate \Cref{equation_of_motion} into two first-order differential equations for $\vec{v}$ and $\vec{x}$ and apply the backward Euler method \cite{baraff_witkin_large_steps}:
%By separating \Cref{equation_of_motion} into two first-order differential equations and applying the backward Euler method \cite{baraff_witkin_large_steps}, we obtain the update scheme
\begin{align}
    \vec{v}_{n+1} & = \vec{v}_n + \Delta t M^{-1} \vec{F}_{n+1} = \vec{v}_n + \Delta t \vec{a}_{n+1} \label{v_update} \\
    \vec{x}_{n+1} & = \vec{x}_n + \Delta t \vec{v}_{n+1} = \vec{x}_n + \Delta t \vec{v}_n + \Delta t^2 \vec{a}_{n+1} \label{x_update}
\end{align}

% mention: scheme is implicit: => stable results / numerical results would be more expensive!
% => thus, we train a neural network ...
Unfortunately, solving this implicit scheme numerically is computationally expensive - even more so, if we want to compute gradients with respect to the solution. 
Semi-implicit or explicit integration schemes are known for instabilities when simulating the stiff equations of cloth \cite{baraff_witkin_large_steps} and lose their benefit in runtime when used with tiny time steps.
Thus, we decided to train a neural network that learns how to solve for the acceleration $\vec{a}_{n+1}$ based on the current positions, velocities, forces, and cloth parameters.
This way, we obtain fast simulations that are naturally differentiable via backpropagation through time.

%For a fast and differentiable solution of \Cref{v_update}, we train a neural network that computes the acceleration $\vec{a}_{n+1}$ based on the forces, masses, current positions and velocities.

\subsubsection{Network Architecture}
\label{sec:network_architecture}

\begin{figure}
    \centering
    \includegraphics[width=0.8\columnwidth]{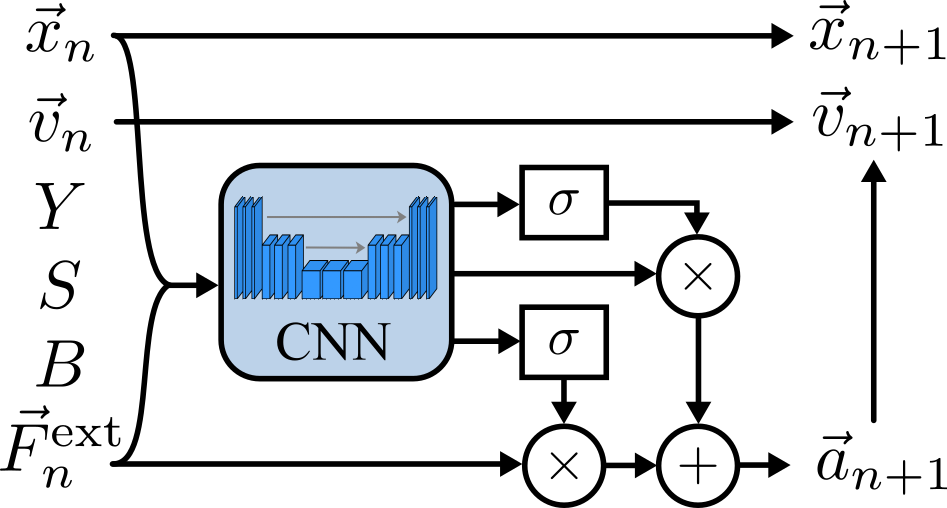}
    \caption{Architecture of the neural cloth model. Detailed explanations are provided in Section \ref{sec:network_architecture}.}
    \label{fig:architecture}
\end{figure}

% overview
The neural cloth model (see Figure \ref{fig:architecture}) is based on a convolutional neural network architecture that maps a rectangular grid of vertex positions $\vec{x}_n$ and velocities $\vec{v}_n$ together with cloth parameters $Y,S,B$ and external forces $\vec{F}_n^\textrm{ext}$ to accelerations $\vec{a}_{n+1}$ in order to update $\vec{v}_{n+1}$ and $\vec{x}_{n+1}$ according to \Cref{v_update,x_update}. 
% CNN based on U-Net (cite ronneberger / implementations from ...)
The CNN block makes use of a U-Net~\cite{unet_ronneberger} implemented by \cite{Yakubovskiy:2019}. 
%gating mechanism
We use a gating mechanism that allows the CNN to directly pass $\vec{F}_n^\textrm{ext}$ to $\vec{a}_{n+1}$ which is useful if the cloth is in free fall. If the cloth hangs in a static equilibrium state, $\vec{a}_{n+1}=0$ and both gates can be closed.

% why CNN and not GNN
The rectangular grid representation of the cloth was chosen since this structure is similar to woven fabric when simulating cloth at yarn-level \cite{Cirio_yarn_1,Gong}. 
%\textcolor{red}{CNN => grid of vertices similar to a woven fabric.}
%This structure is similar to woven fabric when simulating cloth on yarn-level \cite{Cirio_yarn_1,Gong}.
Furthermore, grid computations can be implemented more efficiently on GPUs in comparison to arbitrary graph representations, since no sparse adjacency matrix multiplications are required. 

\paragraph{Normalizations} In order to resolve ambiguities, we introduce special units of measurement in which our network operates.
As the motion of vertices is only affected by the ratio of force and mass, $\vec{a} = M^{-1} \vec{F}$, we fix the mass of interior vertices to be the identity $M = I$ (border and corner vertices are lighter by a factor of $2$ and $4$ respectively).
Moreover, we scale space and time such that edge rest lengths~$L_0^{ij} = 1$ (see \Cref{equation_stretching}) and the simulation time steps~$\Delta t = 1$. %and all other parameters according to these units of measurement.

\subsubsection{Physics-based Training Loss $\texorpdfstring{\mathcal{L}_\textrm{cloth}}{L_cloth}$}

The network learns the dynamics of cloth in a self-supervised manner by minimizing a physics-based loss function $\mathcal{L}_\textrm{cloth}$ similar to \cite{PBNS,SNUG}. %that encodes the cloth's dynamics \cite{PBNS,SNUG}. % encode cloth dynamics finde ich ein bisschen verwirrend
This way, we avoid the need of ground truth data from computationally expensive simulators or other sources. 
The loss function
\begin{equation}
    \mathcal{L}_\mathrm{cloth} = E_\mathrm{int} + \mathcal{L}_\mathrm{ext} + \mathcal{L}_\mathrm{inert} \label{physics_based_loss}
\end{equation}
consists of an internal energy term $E_{\mathrm{int}}$ as well as loss terms that reward accelerations in the direction of external forces $\mathcal{L}_{\mathrm{ext}}$ and penalize sudden changes in momentum $\mathcal{L}_{\mathrm{inert}}$.
%In the following, we describe the individual loss terms in more detail and show that minimizing this loss indeed corresponds to solving the equations of motion.
%In the following, we provide a more detailed description of the individual loss terms and show that minimizing this loss indeed corresponds to solving the equations of motion.

% den folgenden Abschnitt würde ich hinter die loss terme stellen und dann daraus Newtons 2. Bewegungsgleichung herleiten...
% which create forces acting on vertices, i.e. forces for stretching, shearing and bending as well as external forces and an inertia term.
% Note that in every loss term all vertex-related vectors are the updated vectors in order for the network to learn how to improve its returned acceleration.
% We can model most of the forces by setting the corresponding loss contribution equal to a intential because the network minimizes the loss function similar to how nature minimizes intential energy.
% In that case, the derivative of such a Loss term with respect to a network parameter $\alpha$ is 
% \begin{equation}
% \label{derivative_loss}
%     \frac{\partial \mathcal{L}}{\partial \alpha} = \frac{\partial \mathcal{L}}{\partial \vec{x}} \frac{\partial \vec{x}}{\partial \vec{a}} \frac{\partial \vec{a}}{\partial \alpha} = - \vec{F} (\Delta t)^2 \frac{\partial \vec{a}}{\partial \alpha}
% \end{equation}
% which is proportional to the force defined by this intential.
% Since we want to keep everything simple, the forces are modeled using simple intential energies as well although more sophisticated functions are also possible.

\subsubsubsection{Internal Energy $E_\textrm{int}$ and Forces $\texorpdfstring{\vec{F}_\textrm{int}}{F_int}$}
Our cloth model considers a combination
\begin{equation}
    E_\textrm{int} = E_Y + E_S + E_B
\end{equation}
of stretching $E_Y$, shearing $E_S$, and bending $E_B$ components for the total internal energy.
However, different material models can easily be installed by using appropriate energy functions.
The negative gradient of this internal energy term $E_{\textrm{int}}$ with respect to the vertex positions $\vec{x}$ results in forces $\vec{F}_{\textrm{int}}$:
\begin{equation}
    \vec{F}_{\textrm{int}} = -\frac{\partial E_{\textrm{int}}}{\partial \vec{x}}
\end{equation}

\paragraph{Stretching}
To constrain the length of edges $\vec{e}^{\,ij}$ between two neighboring vertices $\vec{x}^i$ and $\vec{x}^j$, we penalize deviations from the rest length $L_0^{ij}$ by a Hookean energy term \cite{PBNS}
\begin{equation}
\label{equation_stretching}
    E_Y = \frac{1}{2} Y \left( \Vert \vec{e}^{\,ij} \Vert - L_0^{ij} \right)^2.
\end{equation}
The constant $Y$ describes the stretching stiffness, i.e. a weight that determines how hard it is for the system to deviate from the minimal energy state.

\begin{figure}
    \centering
    \includegraphics[width=0.97\columnwidth]{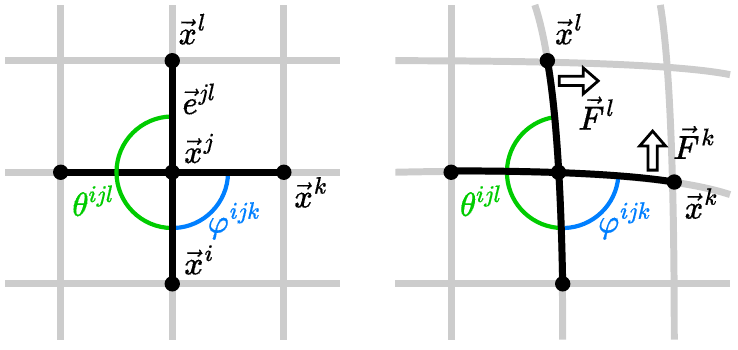}
    \caption{Angles for shearing and bending energies. Straight and corner connections are treated independently to separate in-plane and out-of-plane deformations.}
    \label{fig:e_int_contributions}
\end{figure}

\paragraph{Shearing}
Resistance against angular displacements is taken into account by shearing and bending forces acting on the angle between neighboring edges.
A potential can be modeled using the squared difference between the current angle and the target angle \cite{Curvature_Sullivan2008,Cirio_yarn_1}.
The in-plane angles are restricted by the shearing force between a pair of edges $(\vec{e}^{\,ik}, \vec{e}^{\,kj})$, one of which is pointing in warp direction and the other one in weft direction (see \Cref{fig:e_int_contributions}).
For an enclosed angle $\varphi^{ijk}$, the energy
\begin{equation}
    E_S = \frac{1}{2} S \left( \varphi^{ijk} - \varphi_0^{ijk} \right)^2
\end{equation}
is scaled by the shearing stiffness $S$ and the target angle $\varphi_0^{ijk}$ can be set to $\pi/2$ for a planar fabric \cite{Cirio_yarn_1}.

\paragraph{Bending}
A bending loss is calculated analogously for two consecutive edges with both pointing in the same direction (either warp or weft) as depicted in \Cref{fig:e_int_contributions}.
Nonetheless, a bending stiffness $B$ is used to decouple the optimization of both terms such that the bending energy reads
\begin{equation}
    E_B = \frac{1}{2} B \left( \theta^{ikl} - \theta_0^{ikl} \right)^2
\end{equation}
with rest angles $\theta_0^{ijk}$ \cite{SNUG,Curvature_Sullivan2008}.
In the case of a planar fabric we have $\theta_0^{ijk} \in \{0, \pi\}$, depending on the edge directions.
We assume the parameters $(Y, S, B)$ to be the same for all edges and angles in the cloth.

\subsubsubsection{External Forces $\texorpdfstring{\vec{F}_\textrm{ext}}{F_ext}$}
An arbitrary force field $\vec{F}_\mathrm{ext}$ might not be conservative and, thus, cannot be modeled based on the gradient of a potential.
However, we can define a loss term $\mathcal{L}_{\mathrm{ext}}$ as
\begin{equation}
    \mathcal{L}_{\mathrm{ext}} = - \Delta t^2 \langle \vec{F}_{\mathrm{ext}}, \vec{a} \rangle.
\end{equation}
$\vec{F}_\mathrm{ext}$ can be seen as the effective external forces that act on the grid nodes and result from the superposition of gravity, wind and other forces and may vary over space and time. 
In principle it is possible for these force vectors to include all internal forces as well but this highly reduces the regularization such that learning almost arbitrary realistic dynamics becomes significantly harder \cite{Phi-SfT}. % etwas unklar beim ersten lesen... vielleicht könnte man sagen, dass das von der implementierung (und auch vom rechenaufwand?) komplexer wäre, als wenn man das als potential aufschriebt? evtl könnte man noch sagen, dass das gravitationspotential zwar ebenfalls als potentielle energie beschrieben werden könnte, wir aber die gravitation der einfachheit halber ebenfalls direkt in den external forces mitberücksichtigen...

\subsubsubsection{Inertia}
If the neural network only minimizes the previous loss expressions, the cloth will move to the equilibrium state instantly as no contribution restricts how quickly the vertices move.
To change that, an inertia term of the form
\begin{equation}
    \mathcal{L}_\mathrm{inert} = \frac{1}{2} (\Delta t)^2 \langle \vec{a}, M \vec{a} \rangle 
\end{equation}
is added that penalizes momentum changes of the grid \cite{SNUG}.

\subsubsubsection{Equations of Motion}
By training the neural cloth model, it aims to minimize the loss function $\mathcal{L}_\mathrm{cloth}$ (Equation \ref{physics_based_loss}) causing its gradient with respect to the network's outputs $\vec{a}_{n+1}$ to vanish:
%OLD: By training the neural cloth model on the loss function $\mathcal{L}_\mathrm{cloth}$ (Equation \ref{physics_based_loss}), it aims to find a minimum at which the gradient with respect to the network's outputs $\vec{a}_{n+1}$ go to $0$:
\begin{equation}
    \frac{\partial \mathcal{L}_\mathrm{cloth}}{\partial \Vec{a}_{n+1}} = \underbrace{\frac{\partial E_{\textrm{int}}}{\partial \vec{x}_{n+1}}}_{-\vec{F}_{\textrm{int}}} \underbrace{\frac{\partial \vec{x}_{n+1}}{\partial \vec{a}_{n+1}}}_{\Delta t^2} + \underbrace{\frac{\partial \mathcal{L}_{\textrm{ext}}}{\partial \vec{a}_{n+1}}}_{-\Delta t^2 \vec{F}_{\textrm{ext}}} + \underbrace{\frac{\partial \mathrm{\mathcal{L}}_{\textrm{inert}}}{\partial \vec{a}_{n+1}}}_{+\Delta t^2 M \vec{a}_{n+1}} \overset{!}{=} 0
\end{equation}

Dividing this equation by $\Delta t^2$ leads to an implicit scheme for Newton's second law of motion (Equation \ref{equation_of_motion}):
\begin{equation}
    M \vec{a}_{n+1} = \vec{F}_{\textrm{int}} + \vec{F}_{\textrm{ext}}
\end{equation}
% TODO: link to Equation 1
% mention that this is an implicit step => stable simulation
This way, the neural network can learn stable dynamics of cloth directly from the physics-based loss that ensures physical plausibility and no ground truth data is required.

\subsubsection{Training Cycle}
\label{sec:training_cycle}

\begin{figure}
    \centering
    \includegraphics[width=\columnwidth]{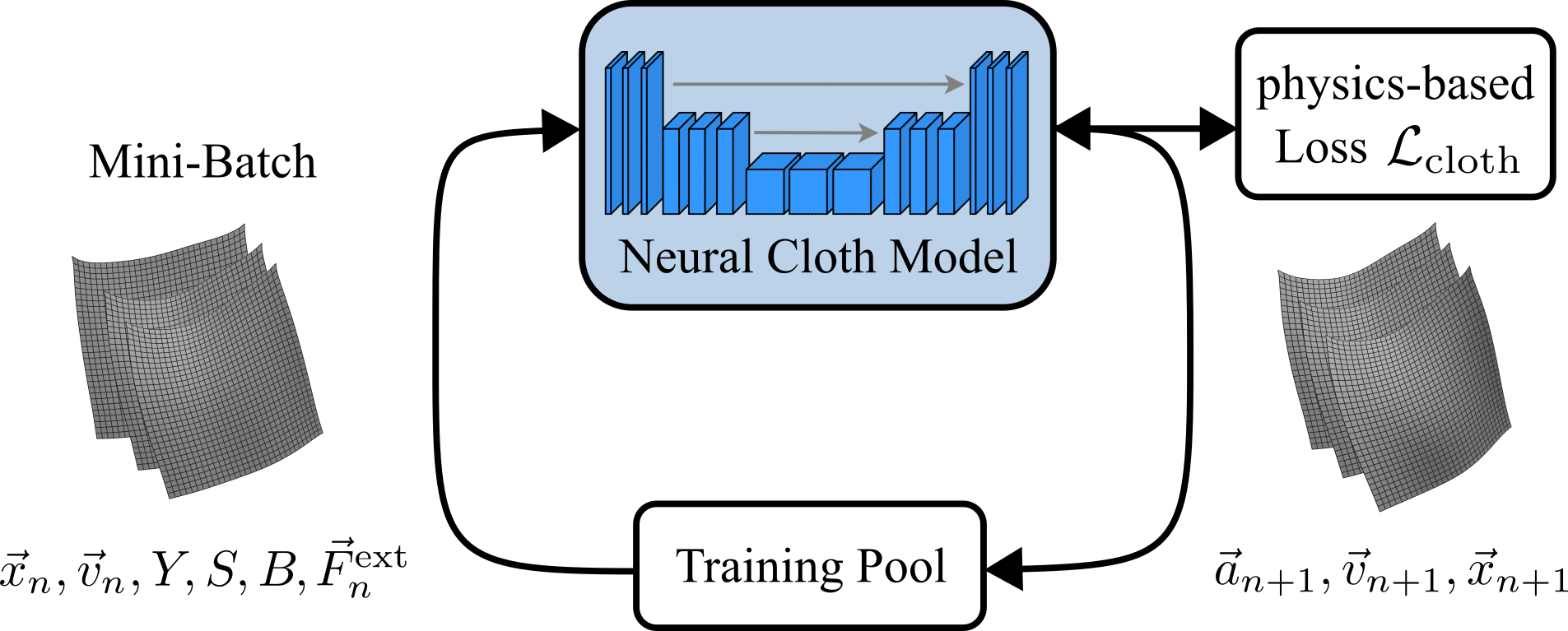}
    \caption{Training cycle of our neural cloth model. Details are provided in Section \ref{sec:training_cycle}.}
    \label{fig:training_cycle}
\end{figure}

To train the neural cloth model, we make use of a training cycle similar to \cite{wandel_2021, HOOD} (see Figure \ref{fig:training_cycle}). %ich zitiere mich hier mal ganz dreist ;)
First, we create a training pool of 5000 cloth states by initializing 32x32 grids at random resting poses ($E_\textrm{int}(\vec{x}_0)=0,\vec{v}_0=0$) with random stretching $Y \in [10,10000]$, shearing $S \in [0.01,10]$ and bending $B \in [0.001, 10]$ parameters.
After that, we draw a random mini-batch of size 300 containing cloth states ($\vec{x}_n,\vec{v}_n$), stiffness parameters ($Y,S,B$), random external forces $\vec{F}_n^\textrm{ext}$ (e.g. gravity and wind), as well as boundary conditions that fasten the top left and right corners of the cloth (see red anchor points in Figure \ref{fig:deformation}). 
Next, we feed the mini-batch into the neural cloth model in order to predict the accelerations $\vec{a}_{n+1}$ for the next timestep. 
This allows to compute $\vec{v}_{n+1}$ and $\vec{x}_{n+1}$ based on \Cref{v_update,x_update} and, thus, to evaluate the physics-based loss (Equation \ref{physics_based_loss}). 
Since $(Y,S,B)$ and therefore $\mathcal{L}_\textrm{cloth}$ can vary drastically, we rescale the loss for every batch sample to equal 1.
We use the Adam optimizer~\cite{adam} (the learning rate starts at $10^{-3}$ and is decreased by $\times 0.2$ after 25 and 50 epochs) to optimize the model.% with gradient descent.
Finally, we feed the predictions of the neural surrogate model back into the training pool to fill it with a larger variety and more realistic training samples. 
If the physics-based loss of a sample becomes too high, we reset the corresponding cloth sample to a new random initial pose to avoid diverged shapes during the training.
Additionally, we randomly reset samples of the training pool from time to time to further increase the variability of training data. 
%\textcolor{red}{TODO: mention:
%random reset or if loss too high (diverged results at the beginning of training)
%loss rescaling
%multiple optimization steps per cycle
%}

By repeating the training cycle for 100 epochs (approx. 10 hours on a Nvidia GeForce RTX 2080), we obtain a fast and stable neural surrogate model to simulate cloth dynamics that can be used for a wide variety of different stiffness, shearing, bending parameters and offers efficient gradient computations based on backpropagation through time.

% TODO: image similar to Phi-Sft pipeline and mention that we replaced cloth simulation by neural cloth model. furthermore, we fix the cloth parameters during (Details for training cloth model are given in Section 3.4 / Details of Differentiable rendering are given in Sec 3.5 ...)

\subsection{Differentiable Rendering}

\subsubsection{RGB-Images and Silhouettes}

Synthetic images are created using the differentiable rasterizing functionality of nvdiffrast~\cite{nvdiffrast}.
This requires vertex positions with $uv$-coordinates and normal vectors, face indices, and a texture file from which everything is known due to an initial template and the physical simulation.
The rasterized image is then used for the optimization step.
Besides the RGB image, a monochromatic mask is saved as well to optimize the visible silhouette, see \Cref{silhouette_loss}.

\subsubsection{Texture Mapping}
\label{section:texture_mapping}
The texture for rendering the cloth is extracted from the first frame of a video sequence which might contain deformations or occlusions caused by folds.
In order to reduce mismatches in the rendered images for later frames, we optimize the $uv$-coordinates of each vertex using the information of all video frames.
We intentionally decided against optimizing the texture directly because imperfect cloth motion will result in blurry textures and contributions of the (black) background.
The effect of this optimization will be analyzed in \Cref{sec:ablation_study}.

\subsection{Shape-from-Template Optimization Loop}

Since each part of our pipeline (see Figure \ref{fig:pipeline}) is differentiable, gradients can be propagated throughout the entire optimization via backpropagation through time. This way, the cloth shape, its physical parameters, and external forces can be estimated by minimizing the difference between rendered and real-world video frames with gradient descent.

\subsubsection{Shape-from-Template Loss}

%\textcolor{red}{Zählt das noch zu "Implementation details" oder sollte man da etwas anders benennen? => diese Kapitel würde ich in Optimization Loss umbennen. Dann zuest high level loss (Full loss) einführen. Dann die einzelnen komponenten des Full losses in den nachfolgenden subsubsubsections erklären... dann wäre das von der Struktur ähnlich zum Neural Cloth Model}

To optimize the shape of the cloth template, we minimize a loss that combines an image loss $\mathcal{L}_\mathrm{im}$, silhouette loss $\mathcal{L}_\mathrm{sil}$ and a regularization term for external forces $R_T$:
\begin{equation}
    \mathcal{L}_\mathrm{SfT} = \mathcal{L}_\mathrm{im} + \mathcal{L}_\mathrm{sil} + R_T
\end{equation}
%is a combination of image and silhouette loss as well as the regularization term.
%Next, we explain the indvidual terms in more detail.

\subsubsubsection{Image Loss}
%The loss function for optimizing the input variables of the simulating network consists of two parts, a RGB image loss $\mathcal{L}_\mathrm{im}$ and a silhouette loss $\mathcal{L}_\mathrm{sil}$.
The image loss averages the pixel-wise differences between the ground truth RGB video frames $\vec{I}_i$ and the generated images $\hat{\vec{I}}_i$ from the simulation:
\begin{equation}
\label{image_loss}
    \mathcal{L}_\mathrm{im} = \frac{1}{N_\mathrm{p}} \sum_{i = 1}^{N_\mathrm{p}} \Vert \hat{\vec{I}}_i - \vec{I}_i \Vert_1
\end{equation}

\subsubsubsection{Silhouette Loss}
As the dataset provides masks for each frame, we also create a silhouette loss.
Therefore, the masks of the ground-truth~$M$ and rendered image~$\hat{M}$ are blurred by a Gaussian kernel~$G$ with standard deviation~$\sigma = 7$ pixels \cite{Phi-SfT} resulting in smooth gradients for the pixel-wise difference.
Again, the corresponding loss function is calculated as the mean of the pixel-wise differences of the blurred masks:% similar to \Cref{image_loss}:
\begin{equation}
\label{silhouette_loss}
    \mathcal{L}_\mathrm{sil} = \frac{1}{N_\mathrm{p}} \sum_{i = 1}^{N_\mathrm{p}} \Vert G(\hat{M}_i,\sigma) - G(M_i,\sigma) \Vert
\end{equation}

\subsubsubsection{Regularization of External Forces}
We separate the external forces $\vec{F}_\mathrm{ext} = \vec{w} + \vec{T}$ into a constant part $\vec{w}$ (e.g. gravity and constant wind) and a spatially and temporarily varying part $\vec{T}$ (e.g. turbulent wind). 
This way, we are able to assign distinct learning rates for $\vec{w}$ and $\vec{T}$ and focus regularization on $\vec{T}$ % evtl kann man diesen Abschnitt auch direkt in sec:reg_T überführen um Platz zu sparen...
to keep external forces stable and smooth. %, we penalize the length of all vectors as well as the spacial and temporal change of them.
To this end, we penalize the turbulent part $T_t^{ij}$ acting at time step $t$ on a vertex with grid indices $i,j$ for its length as well as its temporal and spatial changes:
%This means that for turbulent forces $T_t^{ij}$ acting on vertex $i,j$ at time step $t$ we define the regularization as:
\begin{align}
    \begin{split}
        R_T & = \alpha \sum_{t,i,j} \Vert \vec{T}_t^{ij} \Vert^2 + \beta \sum_{t,i,j} \Vert \vec{T}_{(t+1)}^{ij} - \vec{T}_t^{ij} \Vert^2 \\
        & \hspace{5pt} + \gamma \sum_{t,i,j} \Vert \vec{T}_t^{(i+1)j} - \vec{T}_t^{ij} \Vert^2 + \Vert \vec{T}_t^{i(j+1)} - \vec{T}_t^{ij} \Vert^2
    \end{split}
\end{align}
In our optimization, we set $\alpha = \gamma = 10^{-2}$ and $\beta = 10^{-3}$. % TODO: evtl muss das noch genauer begründet werden!

% \subsubsubsection{Full loss}

% The final optimization loss $\mathcal{L}_\mathrm{op} = \mathcal{L}_\mathrm{im} + \mathcal{L}_\mathrm{sil} + R_\textrm{ext}$ is a combination of image and silhouette loss as well as the regularization term.
% Since each part in this pipeline is differentiable, gradients of this loss function with respect to a physical parameter are computed to update them and minimize the error accordingly.

\subsubsection{Optimization}

Our optimization loop starts with a subset of the first video frames, similar to $\phi$-SfT \cite{Phi-SfT}, and successively adds frames depending on the number of optimization cycles.
This procedure ensures that new dynamics are introduced by new frames only when the previous states match the desired movement.
In our case, we start the optimization with the first 10 frames of a video and add the next frame every 5 optimization cycles until the whole video is considered.

At the beginning of the optimization loop, the model starts with the same physical parameters for all scenes.
The stiffness parameters $(Y, S, B)$ begin at values $(3000, 8, 0.5)$ and their learning rates are set to $(50, 0.1, 0.01)$.
In order to ensure stable simulations, the stiffness parameters are bounded by minimal values $(10, 0.01, 10^{-5})$.
External forces are initialized by the gravitational field $\vec{g}$ only, i.e. $\vec{w} = \vec{g}$, without additional wind or turbulences ($\vec{T} = 0$).
The learning rates are set to $0.05 \cdot \Vert \vec{g} \Vert$ and $0.001$ for constant and turbulent force components respectively.
During optimization, the component along the direction of gravity is fixed such that wind can only appear in horizontal directions.
The $uv$-coordinates are initialized via the known template mesh and updated with a learning rate of $2 \cdot 10^{-4}$.
\section{Evaluation}

We evaluate our method on the $\phi$-SfT dataset \cite{Phi-SfT} which provides several video sequences of real fabrics with diverse movement.
The templates are adjusted to match the requirements of our network, i.e. the geometry is remeshed such that it is represented by a regular $32 \times 32$ grid.
We perform a qualitative analysis of the 3D meshes and compare the rendered images to the target images.
For a quantitative result, we evaluate the precision of the reconstruction as well as the time to converge and compare it with results of $\phi$-SfT~\cite{Phi-SfT}, as it is the state-of-the-art SfT method that uses a physical simulation to regularize the object's movement.
%...as it is the only method that also performs a physics simulation to regularize the cloth movement.
Moreover, we investigate the stability of our algorithm by optimizing for thousands of iterations.
Finally, we perform an ablation study that shows the importance of various features and describe the limitations of our method.

\subsection{Qualitative Evaluation}

\begin{figure}
    \centering
    \includegraphics[width=0.95\columnwidth]{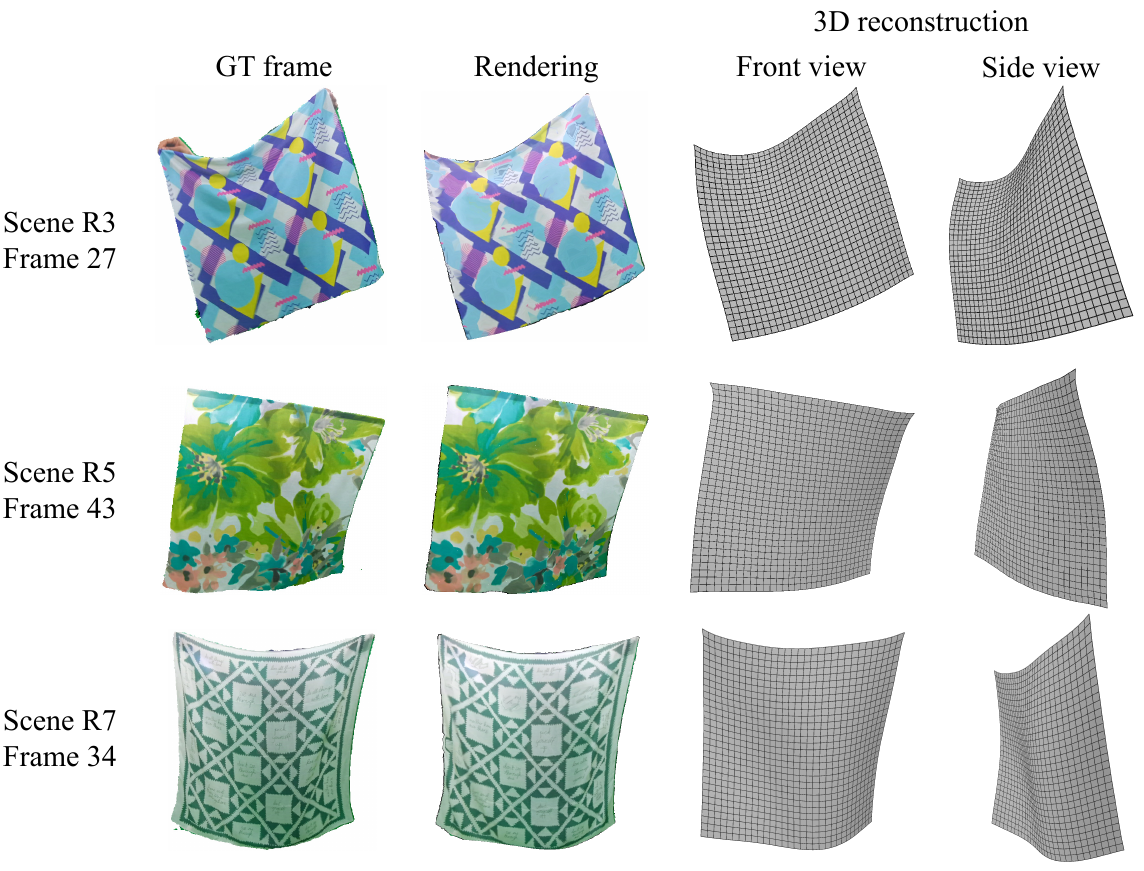}
    \caption{Our approach follows the expected movement in the video and produces smooth 3D geometry for the cloth. Diverse dynamics from manual movement and wind are captured well.}
    \label{fig:qualitative_comparison}
\end{figure}

We evaluate the performance of our method on a dataset of masked images and silhouettes containing several scenes with diverse movements.
First, we look at the reconstructions of our approach qualitatively.
\Cref{fig:qualitative_comparison} depicts the ground truth frame, the rendered image and two views of the reconstructed mesh for late frames in three example scenes.
Generating novel views is no problem due to the physics simulation that generates realistic 3D geometry.
Especially scenes R3 and R5 contain intricate movement due to folding and manual movement.
Nonetheless, our algorithm is able to follow this movement and produce a smooth and stable 3D reconstruction; see our supplementary video for more details.
This behavior is not always the case for SfT and NRSfM methods since assumptions like isometry or conformality do not suffice to prevent the geometry from having an unpleasant and noisy-looking surface \cite{Phi-SfT,DDD,Parashar_2020_CVPR}.
The authors of $\phi$-SfT already showed that other SfT and NRSfM methods usually do not produce smooth and realistic reconstructions and thus a physical simulation is needed to achieve pleasing results \cite{Phi-SfT}.
For this reason, we focus on comparing our method to $\phi$-SfT.

\subsection{Quantitative Evaluation}

We measure the precision and runtime of our method and compare it to the performance of $\phi$-SfT \cite{Phi-SfT}.
All evaluations are performed on an Nvidia A100 GPU and an AMD Epyc 7713 CPU.

The dataset provides different scenes with pseudo ground truth data in form of a target point cloud $T$ per frame.
Therefore, we create a reconstruction point cloud $R$ by sampling the mesh uniformly (each mesh triangle weighted by its area) with the same number of points as the target point cloud.
%In order to measure the precision of the reconstructed mesh, we sample it uniformly (each mesh triangle weighted by its area) with the same number of points as the target point cloud.
The reconstruction quality is then measured by evaluating the symmetric Chamfer distance:
%\begin{align}
%    \begin{split}
%        C(R, T) & = \frac{1}{\vert R \vert} \sum_{\vec{r} \in R} \min_{\vec{t} \in T} \Vert \vec{r} - %\vec{t} \Vert_2^2 \\
%        & \hspace{11pt} + \frac{1}{\vert T \vert} \sum_{\vec{t} \in T} \min_{\vec{r} \in R} \Vert %\vec{r} - \vec{t} \Vert_2^2
%    \end{split}
%\end{align}
\begin{equation}
    C(R, T) = \frac{1}{\vert R \vert} \sum_{\vec{r} \in R} \min_{\vec{t} \in T} \Vert \vec{r} - \vec{t} \Vert_2^2 + \frac{1}{\vert T \vert} \sum_{\vec{t} \in T} \min_{\vec{r} \in R} \Vert \vec{r} - \vec{t} \Vert_2^2
\end{equation}

We use the reference implementation of $\phi$-SfT with default parameters (e.g. 300 optimization cycles) and compare it to our approach after 250 cycles.
Unfortunately, the optimization of scenes R1, R2 and R6 did not finish with this $\phi$-SfT implementation and thus they are discarded for evaluation.
\Cref{tab:quantitative_comparison} shows the evaluated Chamfer distance between the reconstruction and the pseudo ground truth point cloud on real scenes of the $\phi$-SfT dataset.
Both algorithms perform with comparable quality and always produce high-quality results, although $\phi$-SfT matches the target better in 4 out of 6 examples.
Especially in scenes R8 and R9 the results of $\phi$-SfT are very close to the expectations.
%This comes probably from the high-fidelity physics simulation that is able to capture very localized movement better than our physics-based network. 
Nonetheless, our neural model performs better in scenes R5 and R7.

\begin{table}
    \centering
    \setlength{\tabcolsep}{5.6pt}
    \begin{tabular}{lcccccc}
        Method & R3 & R4 & R5 & R7 & R8 & R9 \\
        \midrule
        $\phi$-SfT & \textbf{7.9} & \textbf{10.3} & 14.4 & 9.1 & \textbf{3.7} & \textbf{2.9} \\
        Ours & 12.5 & 14.5 & \textbf{11.7} & \textbf{6.9} & 10.1 & 8.6 \\
        \midrule
        Ratio $\frac{C_\mathrm{Ours}}{C_{\phi\mathrm{-SfT}}}$ & 1.59 & 1.41 & 0.81 & 0.76 & 2.70 & 2.94
    \end{tabular}
    \caption{Quantitative comparison using the symmetric Chamfer distance $C$. All values are multiplied by $10^4$ for readability.}
    \label{tab:quantitative_comparison}
\end{table}

\begin{table}
    \centering
    \setlength{\tabcolsep}{5.9pt}
    \begin{tabular}{lcccccc}
        Method & R3 & R4 & R5 & R7 & R8 & R9 \\
        \midrule
        $\phi$-SfT & 1204 & 1453 & 1152 & 1065 & 1186 & 1157 \\
        Ours & \textbf{3.07} & \textbf{2.48} & \textbf{3.03} & \textbf{2.58} & \textbf{2.55} & \textbf{2.47} \\
        \midrule
        Speedup & 393 & 585 & 380 & 412 & 465 & 469
    \end{tabular}
    \caption{Runtime comparison between $\phi$-SfT \cite{Phi-SfT} and our method. All numbers represent the runtime for the optimization loop until convergence in minutes.
    %The speedup denotes the ratio of our runtimes divided by the runtime of $\phi$-SfT.
    }
    \label{tab:runtime_comparison}
\end{table}

The main goal of our approach and its difference to $\phi$-SfT \cite{Phi-SfT} lies in the runtime.
Due to the neural network, we are able to perform the physics simulation much faster than a classical algorithm could do.
Furthermore, nvdiffrast \cite{nvdiffrast} is significantly faster in rendering images than Pytorch3D.
\Cref{tab:runtime_comparison} shows the runtime of both approaches for all scenes in minutes.
In summary, $\phi$-SfT needs between 17:45\,h and 24:12\,h for optimizing a single scene.
% (depending on the number of frames and the dynamics in the scene).
Compared to that, our method only needs between $2.5$ and $3$ minutes per scene.
This is a speedup of a factor between 380 and 585.
The time for training our physics-based network is not included because this needs to be done only once in advance and the same network is then reused for all scenes.

\subsection{Stability}

\begin{figure}
    \centering
    \includegraphics[width=0.99\columnwidth]{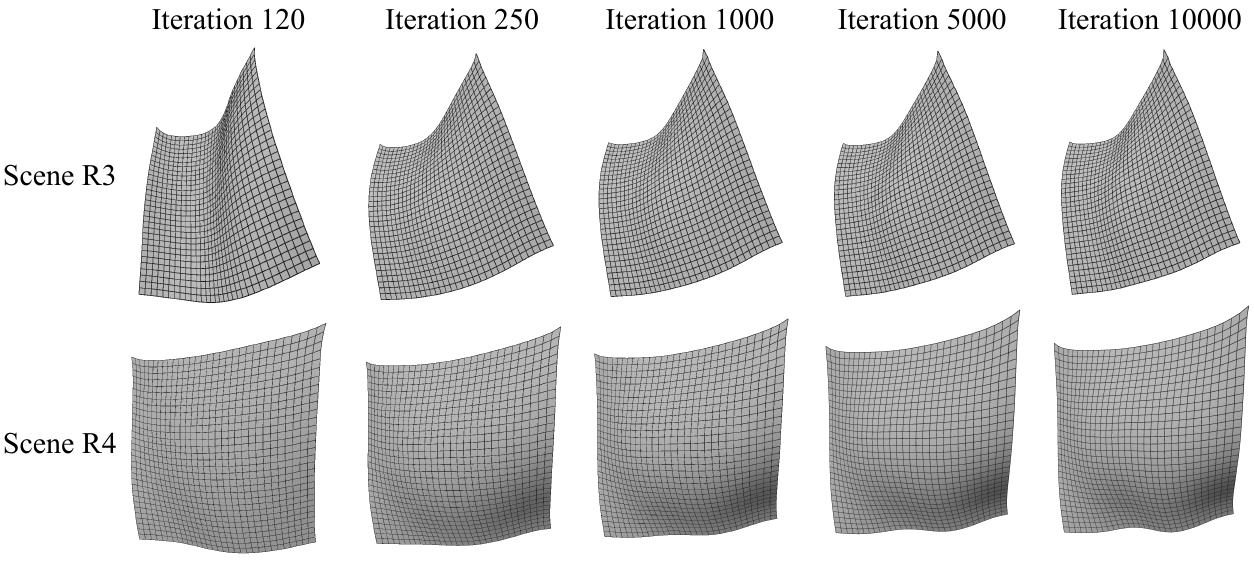}
    \caption{Stability test for R3 and R4 with significant movement. We show a side view of the reconstructed mesh at 120 (before convergence), 250 (regular evaluation), 1000, 5000 and 10000 epochs.}
    \label{fig:stability}
\end{figure}
\begin{figure}
    \centering
    \includegraphics[width=0.99\columnwidth]{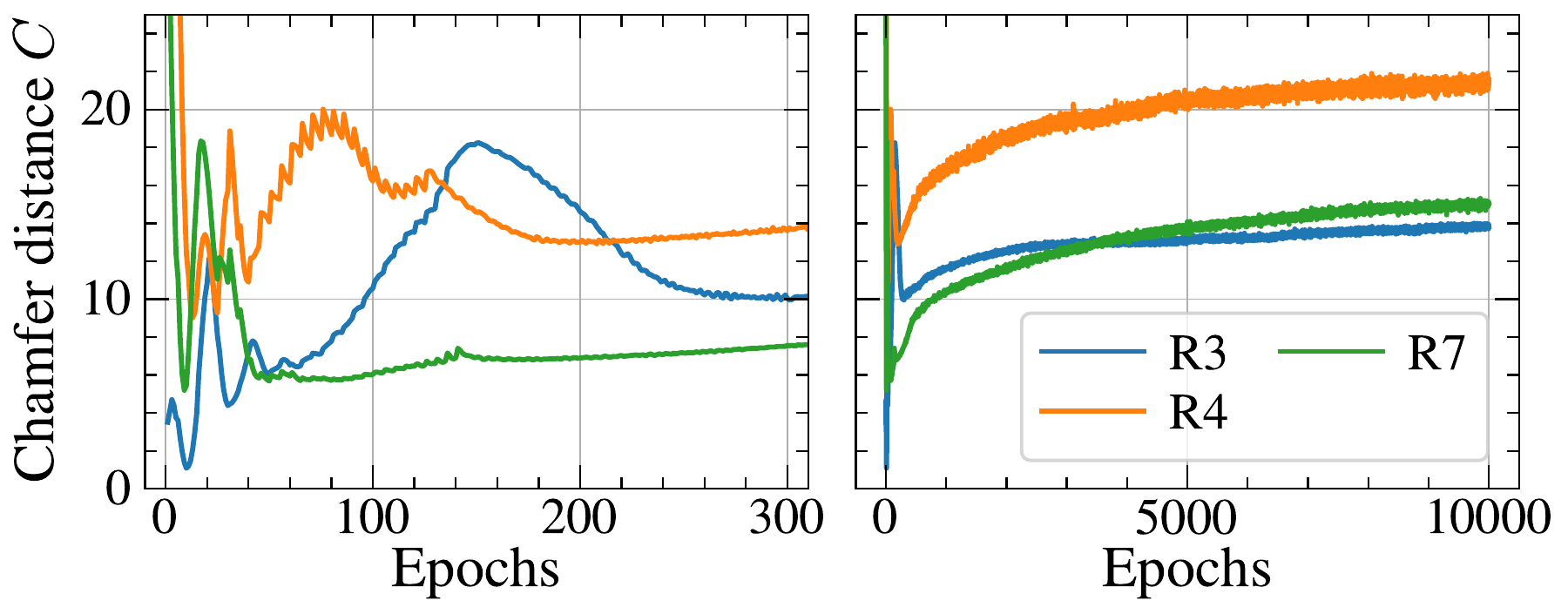}
    \caption{Variation in Chamfer distance $C$ during optimization epochs. After $400$ epochs, overfitting worsens the reconstruction quality but does not affect the stability of the overall optimization. The remaining scenes experience very similar behavior.}
    \label{fig:chamfer_distance_stability}
\end{figure}

We also investigate the stability of our approach by running the optimization loop $40$ times longer than usual.
\Cref{fig:stability} depicts reconstructed meshes from two scenes with diverse movement at different epochs during the optimization loop.
%From left to right the number of iterations increases from 120 iterations (not converged) over 250 (regular evaluation), 1000, 5000 up to 10000 iterations.
It can be seen that our reconstructed mesh still follows the desired movement from the video and does not change significantly after the regular $250$ epochs are reached.
Moreover, \Cref{fig:chamfer_distance_stability} shows how the Chamfer distance $C$ varies during the epochs.
Overfitting worsens the Chamfer distance when our model optimizes for too long but does not affect the stability.
Compared to that, when optimizing for too long, $\phi$-SfT \cite{Phi-SfT} often suffers from instabilities that greatly decrease the reconstruction quality or even lead to a completely different movement than expected.

\subsection{Ablation Study}
\label{sec:ablation_study}

To analyze the importance of each feature in our approach, we perform an ablative study in which we remove one part at a time and compare the results to the full model.
\Cref{tab:ablation_study} comprises variants that miss either the silhouette loss $\mathcal{L}_\mathrm{sil}$, turbulent forces $T^{ij}_t$, their regularization $R_T$, the $uv$-optimization, or the successive optimization scheme.

\begin{table}
    \centering
    \begin{tabular}{lcccccc}
        Ablation & R3 & R4 & R5 & R7 & R8 & R9 \\
        \midrule
        $\mathcal{L}_\mathrm{sil}$ & 13.8 & 14.7 & 12.2 & 7.1 & 10.7 & 8.7 \\
        $R_T$ & 12.6 & 14.8 & 11.9 & 7.0 & 10.4 & \textbf{8.5} \\
        $T_t^{ij}$ & 21.0 & 45.9 & 172.4 & 61.9 & 31.5 & 14.7 \\
        $uv$-map & 13.3 & 24.1 & 14.5 & 61.9 & 20.1 & 30.9 \\
        Suc.-opt. & \textbf{9.7} & 19.0 & 12.4 & 19.0 & 13.8 & 13.6 \\
        Full & 12.5 & \textbf{14.5} & \textbf{11.7} & \textbf{6.9} & \textbf{10.1} & 8.6 \\
    \end{tabular}
    \caption{Reconstruction quality when one feature in our algorithm is removed. Small variations can also occur due to randomness.}
    \label{tab:ablation_study}
\end{table}

The silhouette loss only slightly improves the results in all scenes.
Unsurprisingly, the largest difference is present in scene R3 where one of the two holding points is moving significantly and causing the silhouette to change much more than in other scenes.
%This result shows that it might be possible to drop the assumption of masked images and silhouettes without losing significant amount of quality.
Removing the regularization $R_T$ also does not change the quality significantly within the given number of iterations.
The simulating network itself processes the external forces and regularizes their effect due to convolutional layers but we want to control them explicitly (also regarding the long-time behavior).
However, when turbulent forces are neglected completely, large mismatches are caused.
These turbulent forces are not only used to create local deformations that only affect a few vertices but also for moving the anchor points.
Especially in scene R5 the anchor points create the majority of the dynamics.
The $uv$-map optimization has a large effect on the reconstruction precision because the texture is taken from the first RGB frame.
This might include perspective effects and occlusions that introduce a bias into the rendering (see \Cref{section:texture_mapping}).
%By optimizing the $uv$-map, we use the information of every frame instead of only the first one. %; although the texture file is equal to the first frame of the video sequence.
Finally, we omit the successive optimization scheme, i.e. we optimize with all frames from the beginning.
Again, we observe a decreasing reconstruction quality in all scenes except R3. 
%In R3, the direction in which the cloth bends changes which causes the geometry to be closer to the target.

\subsection{Limitations}

\paragraph{Fine wrinkles}

\begin{figure}
    \centering
    \includegraphics[width=0.96\columnwidth]{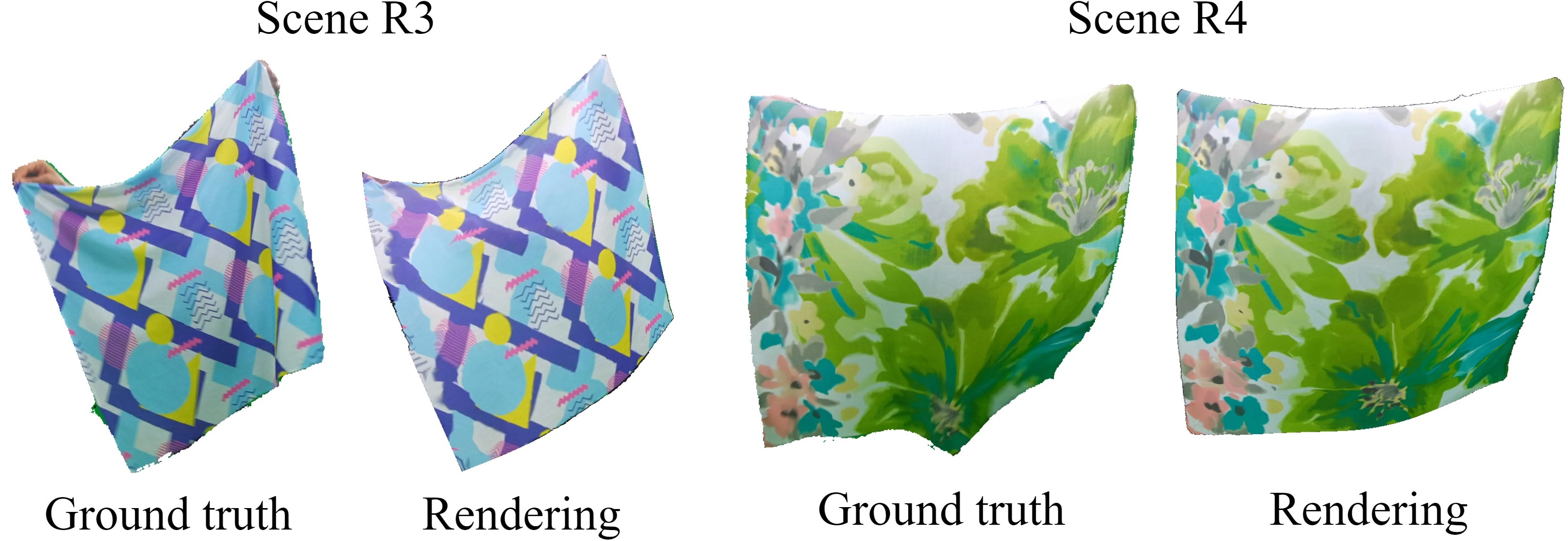}
    \caption{Our network does not capture high-curvature details like sharp folds very well. Such effects occur due to manual movement in scene R3 or wind in R4.}
    \label{fig:fine_folds}
\end{figure}

\Cref{fig:fine_folds} shows examples in which our network is not able to simulate fine wrinkles and reconstruct corresponding details.
This is visible in folds (scene R3) and high-frequency movement from wind turbulences (R4).
Possible solutions are a higher grid resolution or different physical energy terms for shearing and bending forces. 

\paragraph{$uv$-mapping}

\begin{figure}
    \centering
    \includegraphics[width=0.96\columnwidth]{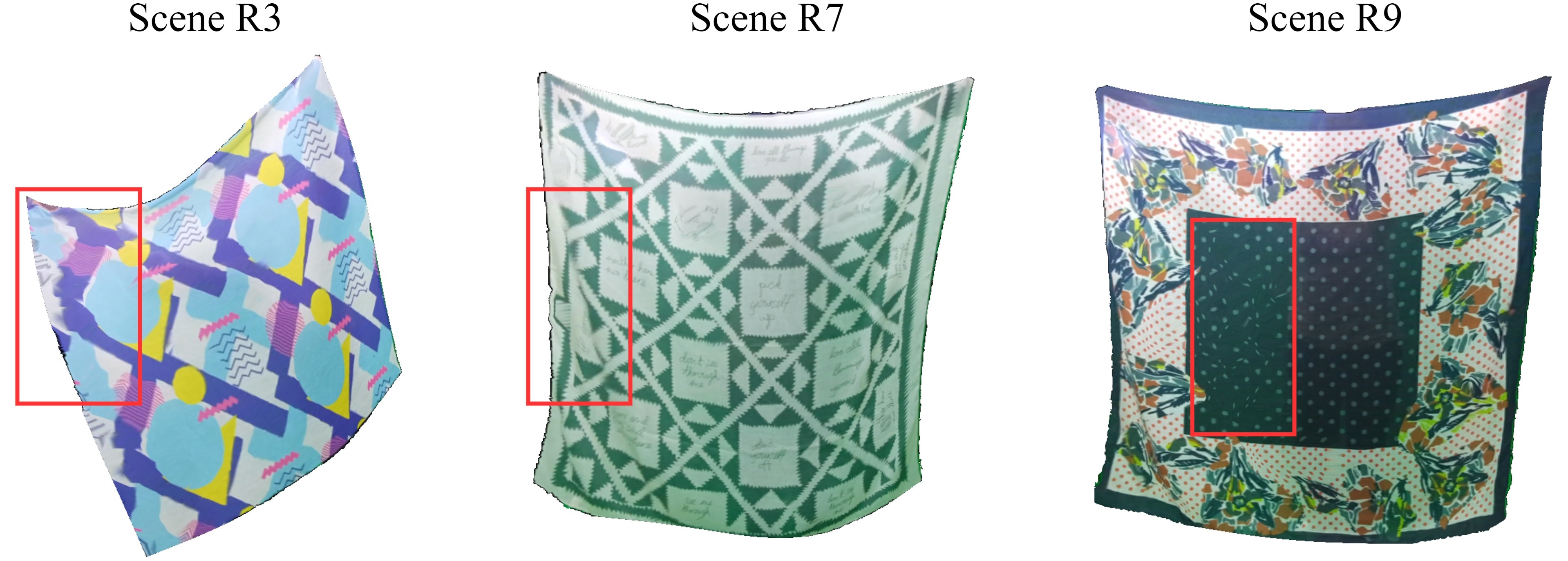}
    \caption{Optimizing the $uv$-map leads to some artifacts (red boxes) in the texture which are most noticeable when the texture contains regular geometric structures like straight lines or dots.}
    \label{fig:uv-map}
\end{figure}

The ablation study shows that optimizing the $uv$-map significantly improves the precision of the reconstructed mesh.
However, the ability of warping the texture by need also opens the possibility of creating distorted texture mappings.
Especially small details and geometric patterns can suffer from that effect.
Three examples are shown in \Cref{fig:uv-map} depicting the optimized cloth of scenes R3, R7 and R9.
The cloth textures in scenes R3 and R7 contain several straight lines that get smeared or shifted due to the movement.
Nevertheless, most parts of the $uv$-map barely changed even without a regularization such that intricate patterns or words (scene R7) are still intact.
These Issues could be solved by regularizing the $uv$-coordinates or by optimizing the texture directly.
Other methods do not optimize the $uv$-map at all and keep potential deformations from the first frame which leads to worse reconstructions.

\section{Conclusion}

We presented a novel Shape-from-Template method that reconstructs the 3D geometry of a piece of cloth together with physical parameters for stretching, shearing, and bending based on a single monocular RGB video sequence and a template mesh.
We employ a physics-based neural network that enables fast and stable physical simulation without the need of costly classical simulation methods.
This simulation regularizes possible dynamics of the 3D geometry for the optimization process.
Our method was compared to the state-of-the-art physics-based $\phi$-SfT method and achieved comparable results with a speedup of 400--500 times.

%We presented a novel Shape-from-Template method that reconstructs the 3D geometry of a piece of cloth based on a single monocular RGB-video sequence and a template mesh.
% The key idea is the movement regularization by a physics simulation for which we employ a physics-based neural network.
%This leads to realistic deformations without the need of a sophisticated classical simulation.
%A differentiable rendering algorithm creates images and masks that are compared to the target video frames.
%A loss function calculated from the difference images allows for a gradient-based update procedure of the input parameters due to the differentiability of the whole approach.
%We analyzed our method with respect to qualitative and quantitative quality as well as the time needed for optimizing the dynamics.
%These results are compared to the state-of-the-art SfT method that also uses a physics simulation to regularize the cloth's movement.
%Summarizing, we are able to reconstruct the desired geometry around 400--500 times faster than the existing approach while keeping comparable quality.
%Moreover, we showed the stability of our algorithm by running our optimization loop several times longer than necessary.
%Finally, we point out inaccuracies and improvements to our method and discuss possible extensions.
\section*{Acknowledgements}

This work has been funded by the DFG project KL 1142/11-2 (DFG Research Unit FOR 2535 Anticipating Human Behavior), by the Federal Ministry of Education and Research of Germany and the state of North-Rhine Westphalia as part of the Lamarr-Institute for Machine Learning and Artificial Intelligence and the InVirtuo 4.0 project, and additionally by the Federal Ministry of Education and Research under grant no. 01IS22094E WEST-AI.
{
    \small
    \bibliographystyle{ieeenat_fullname}
    \bibliography{main}
}


\end{document}